\documentclass{article}

\usepackage{arxiv}

\usepackage[utf8]{inputenc} 
\usepackage[T1]{fontenc}    
\usepackage{hyperref}       
\usepackage{url}            
\usepackage{booktabs}       
\usepackage{amsfonts}       
\usepackage{nicefrac}       
\usepackage{microtype}      
\usepackage{multirow}
\usepackage{graphicx}
\usepackage{amsmath}
\usepackage{amssymb}
\usepackage{algorithm}
\usepackage{algpseudocode}
\usepackage[caption=false]{subfig}

\title{Evolving Gaussian Process kernels \\ from elementary mathematical expressions}

\author{
Ibai Roman\\
Intelligent Systems Group\\
University of the Basque Country UPV/EHU\\
20018 Donostia, Spain\\
\texttt{ibai.roman@ehu.eus}\\
   \And
Roberto Santana\\
Intelligent Systems Group\\
University of the Basque Country UPV/EHU\\
20018 Donostia, Spain\\
\texttt{roberto.santana@ehu.eus}\\
   \And
Alexander Mendiburu\\
Intelligent Systems Group\\
University of the Basque Country UPV/EHU\\
20018 Donostia, Spain\\
\texttt{alexander.mendiburu@ehu.eus}\\
   \And
Jose A. Lozano\\
Intelligent Systems Group\\
University of the Basque Country UPV/EHU\\
20018 Donostia, Spain\\
Basque Center for Applied Mathematics BCAM\\
48009 Bilbao, Spain\\
\texttt{ja.lozano@ehu.eus}\\
}

\begin{document}
\maketitle

\begin{abstract}
Choosing the most adequate kernel is crucial in many Machine Learning applications. Gaussian Process is a state-of-the-art technique for regression and classification that heavily relies on a kernel function. However, in the Gaussian Process literature, kernels have usually been either ad hoc designed, selected from a predefined set, or searched for in a space of compositions of kernels which have been defined a priori. In this paper, we propose a Genetic-Programming algorithm that represents a kernel function as a tree of elementary mathematical expressions. By means of this representation, a wider set of kernels can be modeled, where potentially better solutions can be found, although new challenges also arise. The proposed algorithm is able to overcome these difficulties and find kernels that accurately model the characteristics of the data. This method has been tested in several real-world time-series extrapolation problems, improving the state-of-the-art results while reducing the complexity of the kernels.
\end{abstract}

\keywords{Evolutionary search \and Gaussian Processes \and Genetic-Programming \and Kernel Selection \and Time-series extrapolation}

\section{Introduction}\label{sec:intro}

Gaussian Processes (GPs) \cite{rasmussen_gaussian_2006} are one of the most used techniques in Machine Learning for regression and classification tasks. Furthermore, they have also been applied to optimization tasks under the umbrella of Bayesian optimization \cite{mockus_application_1978}. A GP is a collection of random variables, any finite set of which has a joint Gaussian distribution. It is completely defined by a mean function and a covariance function described in terms of a Positive Semi-Definite (PSD) kernel. The assumption in a GP is that, as the similarity between two solutions increases, so does the similarity of the output function at these solutions. The kernel function encodes the particular manner in which the similarity between any two solutions is defined, which makes it a key element in any application of GPs.
 
While there is a repertoire of kernel functions available in the literature \cite{rasmussen_gaussian_2006,duvenaud_automatic_2014,genton_classes_2002}, the choice of the most appropriate kernel for a given problem is not straightforward. Moreover, kernels usually have some parameters that need to be adjusted, which hardens the kernel selection problem. These parameters, often called hyperparameters, are usually tuned by maximizing a given metric (e.g., the marginal likelihood) \cite{blum_optimization_2013}.
 
In early applications of GPs, the kernel function used to be designed by an expert \cite{rasmussen_gaussian_2006}, or selected from a predefined set \cite{brochu_tutorial_2010}. However, some recent works tackle the question of automating the choice of the kernel \cite{lloyd_automatic_2014, kronberger_evolution_2013, duvenaud_structure_2013}. Compositional kernel search is one of the most used techniques when automating the kernel choice. In this technique, the kernel is always the combination of a limited number of a priori defined kernels, and the kernel selection is reframed as a search in the space of possible kernel compositions. The compositional kernel search approaches take advantage of some operands (e.g., sum, product, ...) that guarantee the composed kernel is PSD as long as its components are also PSD. For example, in \cite{lloyd_automatic_2014} and \cite{duvenaud_structure_2013}, the kernel search is carried out by means of a greedy search procedure. A similar approach is presented in \cite{kronberger_evolution_2013}, although in this work the search is guided by Genetic-Programming (GenProg\footnote{Note that we keep the acronym GP to refer to Gaussian Process, and GenProg to refer to Genetic-Programming.}) \cite{koza_genetic_1992}. The solutions proposed by these methods have shown their ability to capture function properties such as smoothness, trends and periodicity \cite{lloyd_automatic_2014,duvenaud_structure_2013}. In addition, as the behavior of the base kernels and the operands is well-known, the behavior of their composition may be guessed by an expert \cite{lloyd_automatic_2014}. On the contrary, these kernel search methods usually end up with very complicated kernels, including many hyperparameters, which make them expensive to optimize. Moreover, it must be noted that these approaches rely on kernels that have already been proposed in the literature. There is no reason to believe that kernels obtained by composing a limited set of human-designed kernels are optimal for arbitrary problems. Furthermore, using previously designed kernels as building blocks could bias the search and prevent the exploration of more promising candidates.

In this paper, instead of considering a reduced set of base kernels as building blocks, we propose using a set of elementary mathematical expressions (e.g., product, sum, exponent, etc.) to serve as components of a wider set of kernels. Our hypothesis is that, by enlarging the space of possible solutions, the representation capability of GPs is expanded, allowing more accurate kernels with a lower number of hyperparameters to be found.

Searching for the most appropriate solution in the space of mathematical expressions is very challenging due to the vast number of kernels that can be generated and the lack of guarantee that these kernels satisfy the PSD property. Note that, before evaluating a kernel, its hyperparameters should be optimized, which limits the number of kernels that can be explored due to the computational effort required to find these hyperparameters. We propose a novel GenProg method, EvoCov, which is able to overcome these challenges and learn adequate kernel functions for each problem. This method does not rely on previously proposed kernels, and thus, new kernels may naturally arise.

Although we focus on regression problems in this work, our contribution can be easily extended to other GP applications, such as classification. Moreover, some of the components designed in EvoCov could be extended to other GenProg applications. 

The remainder of the paper is structured as follows: In the next section, a background on GP regression is provided, including the presentation of the best known GP kernel functions. In Section~\ref{sec:newgrammar}, we present our kernel representation approach, based on elementary mathematical expressions. In Section~\ref{sec:kernelsearch}, a novel GenProg method, EvoCov, is proposed to search for GP kernel functions based on such grammar. In Section~\ref{sec:relwork}, a review on related work is provided. Next, in Section~\ref{sec:experiments}, we present an empirical validation of the algorithm and comparisons with other methods. Finally, in Section~\ref{sec:conclu}, the conclusions and the future work are presented.


\section{Gaussian Process Regression}
\label{sec:gpr}
A GP is a stochastic process, defined by a collection of random variables, any finite number of which have a joint Gaussian distribution \cite{rasmussen_gaussian_2006}. A GP can be interpreted as a distribution over functions, and each sample of a GP as a function.

GPs can be completely defined by a mean function $m(\mathbf{x})$ and a covariance function, which depends on a PSD kernel $k(\mathbf{x},\mathbf{x}')$. Given that, a GP can be expressed as follows: 

\begin{equation} \label{eq:gp}
f(\mathbf{x}) \sim GP(m(\mathbf{x}), k(\mathbf{x},\mathbf{x}'))
\end{equation}
where we assume that $\mathbf{x} \in \mathbb{R}^d$.

A GP can be used for regression by getting its posterior distribution given some (training) data. The joint distribution between the training outputs $\mathbf{f}=(f_1,f_2,...,f_n)$ (where $f_i \in \mathbb{R}$, $i \in \{ 1,...,n \}$ and $n\in\mathbb{N}$) and the test outputs $\mathbf{f}_{*}=(f_{n+1},f_{n+2},...,f_{n+n_{*}})$ is given by:

\begin{equation} \label{eq:joint}
\begin{bmatrix}
\mathbf{f} \\ \mathbf{f}_{*}
\end{bmatrix} \sim \mathcal{N} \left( M(X_{*}),
\begin{bmatrix}
K(X, X) & K(X, X_{*}) \\ K(X_{*}, X) & K(X_{*}, X_{*})
\end{bmatrix} \right)
\end{equation}
where $N(\mu,\Sigma)$ is a multivariate Gaussian distribution, $X=(\mathbf{x}_1,\mathbf{x}_2,...,\mathbf{x}_n)$ ($\mathbf{x}_i \in \mathbb{R}^d$, $i \in \{ 1,...,n \}$  and $n\in\mathbb{N}$) corresponds to the training inputs and $X_{*}=(\mathbf{x}_{n+1},\mathbf{x}_{n+2},...,\mathbf{x}_{n+n_{*}})$ to the test inputs. $K(X, X_{*} )$ denotes the $n \times n_{*}$ matrix of the covariances evaluated for all the $(X, X_{*})$ pairs.

The predictive Gaussian distribution can be found by obtaining the conditional distribution given the training data and the test inputs:
\begin{equation} \label{eq:gpr}
\begin{split}
\mathbf{f}_{*} | X_{*}, X, \mathbf{f} &\sim \mathcal{N}(\Hat{M}(X_{*}), \Hat{K}(X_{*}, X_{*}))\\
\Hat{M}(X_{*}) &= M(X_{*}) + K(X_{*}, X)K(X, X)^{-1} \mathbf{f} \\
\Hat{K}(X_{*}, X_{*}) &= K(X_{*}, X_{*}) - K(X_{*}, X)K(X, X)^{-1}K(X, X_{*})
\end{split}
\end{equation}

As in many previous works \cite{chu_gaussian_2005,brochu_tutorial_2010,wang_theoretical_2014}, we consider an a priori equal-to-zero mean function ($m(\mathbf{x})=0$) to focus on the kernel search problem.

\subsection{Covariance function}
\label{ssec:kernel}

GP models use a PSD kernel to define the covariance between any two function values \cite{duvenaud_automatic_2014}:

\begin{equation} \label{eq:kernel}
cov\left( f(\mathbf{x}), f(\mathbf{x}') \right) = k(\mathbf{x}, \mathbf{x}')
\end{equation}


A PSD kernel is a symmetric function $k: \mathbb{R}^d \times \mathbb{R}^d \rightarrow \mathbb{R}$ that satisfies
\begin{equation} \label{eq:psdkernel}
\sum_{i=1}^{n} \sum_{j=1}^{n} a_i a_j k(\mathbf{x}_i, \mathbf{x}_j) \geq 0
\end{equation}
for any $n \in \mathbb{N}$,  $\mathbf{x}_1,...,\mathbf{x}_n\in\mathbb{R}^d$ and $a_1,...,a_n\in\mathbb{R}$.

If a kernel function holds Equation~\eqref{eq:psdkernel}, then the covariance matrix $C$, where $c_{ij}=k(\mathbf{x}_i,\mathbf{x}_j)$, $n\in\mathbb{N}$ and $\mathbf{x}_1,...,\mathbf{x}_n \in  \mathbb{R}^d$ is:
\begin{itemize}
  \item symmetric, i.e., $C=C^T$. 
  \item a PSD matrix. A matrix is PSD if and only if $\mathbf{v}^T C \mathbf{v} \geq 0$ for all real vectors $\mathbf{v} \in \mathbb{R}^n$. It is equivalent to say that all its eigenvalues are non-negative.
\end{itemize}

\subsection{Well-known kernel functions}
\label{ssec:wellknownkernels}

In this section, we introduce some of the best-known kernel functions. These kernels can be divided into two main families: stationary and non-stationary kernels \cite{genton_classes_2002}.


A stationary kernel is translation invariant. Among the stationary kernels, we focus on isotropic kernels, as they are the most used kernel functions in the literature. In such kernels, the covariance function depends on the norm:
\begin{equation} \label{eq:stationary}
k(\mathbf{x}, \mathbf{x}') = \hat{k}(r) \text{  where  } r= \frac{1}{\theta_{l}} \left\lVert\mathbf{x}-\mathbf{x}'\right\rVert
\end{equation}
where $\theta_{l}$ is the lengthscale hyperparameter and $\hat{k}$ a function that guarantees that the kernel is PSD. The lengthscale hyperparameter can be also a vector that expresses the relevance of each dimension $d$, as suggested in Automatic Relevance Determination (ARD) approaches \cite{mackay_bayesian_1996,neal_bayesian_1996}.


On the other hand, non-stationary kernels are the ones that may vary with translation. Within this family, the most common kernels are those that depend on the dot product of the input vectors. These kernels are usually referred to as dot-product kernels:
\begin{equation} \label{eq:dotproduct}
k(\mathbf{x}, \mathbf{x}') = \hat{k}(s) \text{  where  } s=\frac{1}{\theta_{l}}\left(\mathbf{x} - \theta_{s} \mathbf{1} \right) \left(\mathbf{x}' - \theta_{s} \mathbf{1} \right)^T
\end{equation}
where $\theta_{l}$ is again the lengthscale hyperparameter, $\theta_{s}$ is the shift hyperparameter and $\mathbf{1}$ is a vector of ones.

Table \ref{tab:kern} shows eleven well-known kernels generally used in GP applications \cite{rasmussen_gaussian_2006,duvenaud_automatic_2014}. The Squared Exponential (SE) kernel, also known as radial basis function (RBF), is one of the most popular choices, and it is described as $k_{SE}$ in the table. This kernel is known to capture the smoothness property of the objective functions.

\begin{table}[!ht]
  \centering
  \def\arraystretch{1.2}
  \begin{tabular}{|l|l|}
    \hline
    \multicolumn{2}{|c|}{Kernel function expressions} \\
    \hline
    Constant & $k_{CON} (\mathbf{x}, \mathbf{x}') = \theta_{c} $\\
    White Noise & $k_{WN} (\mathbf{x}, \mathbf{x}') = \theta_{c} \ \delta(\mathbf{x}, \mathbf{x}') $\\
    Exponential & $\hat{k}_{E} (r) = \theta_{0}^2 \ exp \left( - r\right) $\\
    $\gamma$-exponential &  $\hat{k}_{E\gamma} (r) = \theta_{0}^2 \ exp \left( - r^{\gamma} \right)$ \\
    Squared Exp. &  $\hat{k}_{SE} (r) = \theta_{0}^2 \ exp \left( -\frac{1}{2} r^2 \right)$  \\
    Matern $12$ &  $\hat{k}_{M12} (r) = \theta_{0}^2 exp \left( - r \right)$  \\
    Matern $32$ &  $\hat{k}_{M32} (r) = \theta_{0}^2 \left( 1 + \sqrt{3} r \right) exp \left( - \sqrt{3} r \right)$  \\
    Matern $52$ &  $\hat{k}_{M52} (r) = \theta_{0}^2 \left( 1 + \sqrt{5} r + \frac{5}{3} r^2 \right) exp \left( -\sqrt{5} r \right) $  \\
    Rat. Quadratic &  $\hat{k}_{RQ} (r) = \theta_{0}^2 \left(1+ \frac{1}{2\alpha} r^2 \right)^{-\alpha}$ \\
    Periodic &  $\hat{k}_{PER} (r) = \theta_{0}^2 \exp\left(-\frac{2\sin^2(\pi r)}{\theta_{p}^2}\right) $\\
    Linear &  $\hat{k}_{LIN} (s) = s $\\
    \hline
  \end{tabular}
  \vspace{0.15cm}
  \caption{Well-known kernel functions. $\theta_{0}$ and $\theta_{p}$ are the kernel hyperparameters called amplitude and period respectively. $\delta$ is the Kronecker delta. }
  \label{tab:kern}
\end{table}

\subsection{Model selection}
\label{ssec:modelsel}

The choice of the kernel function and its hyperparameters has a critical influence on the behavior of the model, and it is crucial to achieve good results in any application of GPs. This selection has been usually made by choosing one kernel a priori, and then adjusting the hyperparameters of the kernel function so to optimize a given metric for the data.

 



Although a variety of methods have also been proposed to optimize the hyperparameters \cite{sundararajan_predictive_2001,toal_kriging_2008,garnett_active_2014,toal_development_2011}, the most common approach is to find the hyperparameter set that maximizes the log marginal likelihood (LML):
\begin{equation} \label{eq:lml}
\begin{split}
log \: p \left( \mathbf{f}| X, \boldsymbol\theta \right) &= -\frac{1}{2} m_a^T K_a^{-1} m_a - \frac{1}{2} log \: |K_a| - \frac{n}{2} \: log \: 2\pi \\
with \\
m_a &= \mathbf{f} - M(X) \\
K_a &= K(X,X)
\end{split}
\end{equation}
where $\boldsymbol\theta$ is the set of hyperparameters of the kernel and $n$ is the length of $X$.

Alternatively, a leave-one-out cross validation (LOOCV) metric was proposed \cite{rasmussen_gaussian_2006}, where the likelihood of the posterior distributions are added:
\begin{equation} \label{eq:loocv}
\begin{split}
L_{LOO}(X, \mathbf{f}, \boldsymbol\theta) &= \sum_{i=1}^{n} log \: p \left( f_i| X, \mathbf{f}_{-i}, \boldsymbol\theta \right)
\\
log \: p \left( f_i| X, \mathbf{f}_{-i}, \boldsymbol\theta \right) &= -\frac{(f_i-\mu_i)^2}{2\sigma_i^2} - \frac{1}{2} log \: \sigma_i^2 - \frac{1}{2} \: log \: 2\pi
\end{split}
\end{equation}
where $\mu_i$ and $\sigma_i$ are the posterior mean and variance for $\mathbf{x}_{i}$ given $X_{-i}$ and $\mathbf{f}_{-i}$ \footnote{$X_{-i}$ notation is used to indicate that $\mathbf{x}_{i}$ was removed from $X$.}.

The selection of the right set of hyperparameters is known to be a hard problem, particularly when few observations are available \cite{wang_theoretical_2014,benassi_robust_2011,bull_convergence_2011}. Although in most cases the gradient of the LML and the LOOCV has a closed-form expression, depending on the problem, these functions can be multi-modal and a greedy search procedure may lead to suboptimal results.


\subsection{Kernel composition}
\label{ssec:kernelcomp}

When creating new kernels, it is usually difficult to prove whether they are PSD or not. However, there are certain operations which guarantee that if the source kernels are PSD, the result is also a PSD kernel \cite{duvenaud_automatic_2014,durrande_additive_2012}. Below we list some of the operations that are guaranteed to keep the positive semi-definiteness of its components:

\begin{itemize}
  \item Sum: $k(\mathbf{x}, \mathbf{x}') = k_1 (\mathbf{x}, \mathbf{x}') + k_2 (\mathbf{x}, \mathbf{x}')$.
  \item Product: $k(\mathbf{x}, \mathbf{x}') = k_1 (\mathbf{x}, \mathbf{x}') \times k_2 (\mathbf{x}, \mathbf{x}')$.
  \item Polynomial: $k(\mathbf{x}, \mathbf{x}') = p (k_1 (\mathbf{x}, \mathbf{x}'))$, where $p$ is a polynomial function with non-negative coefficients.
  \item Exponential: $k(\mathbf{x}, \mathbf{x}') = exp(k_1 (\mathbf{x}, \mathbf{x}'))$.
  \item Composition with a function: $k(\mathbf{x}, \mathbf{x}') = f(\mathbf{x}) k_1 (\mathbf{x}, \mathbf{x}') f(\mathbf{x}')$, with $f:\mathbb{R}^d\rightarrow \mathbb{R}$.
  \item Mapping: $k(\mathbf{x}, \mathbf{x}') = k_1 (\Psi(\mathbf{x}), \Psi(\mathbf{x}'))$, with $\Psi:\mathbb{R}^d\rightarrow \mathbb{R}^d$.
\end{itemize}




\section{Gaussian Process kernel representation as elementary mathematical expression trees}
\label{sec:newgrammar}

While previous approaches have proposed the composition of kernel functions, in this work we break down the well-known kernels of Table \ref{tab:kern} into basic mathematical expressions, in order to use them as the building blocks for new kernels. Thus, these well-known kernel functions and their sum/product compositions are a subset of our search space.


The kernel functions are described as an expression tree, composed by the set of the operators and terminals shown in Table~\ref{tab:grammar}. The expression tree is strongly-typed \cite{montana_strongly_1995}, as the output of each node matches the input type of its ancestor.

Our grammar considers a pair of input vectors (their symbol is denoted as $x$) and a vector of hyperparameters (denoted as $hp1, hp2, ...$) as arguments of the kernel function. As in \cite{hajighassemi_analytic_2014}, the grammar also contains the spectral transformation, in order to allow periodic kernels. Furthermore, the square distance and dot product are included as described in Section~\ref{ssec:kernel}. Note that the expressions denoted by the symbols \textit{euc} and \textit{hp} do not modify the input. These expressions are included to constrain the random kernel generation as explained in detail in Section~\ref{sssec:randgen}. In addition, our grammar includes the $+$, $\times$, and \^{} arithmetic operators, having their usual meanings (addition, product and power, respectively). Note that we only allow hyperparameters as the exponent in the power operator. The same interpretation is given to the unary operators, such as the square root and the natural exponent. The power to the minus one is also added as an unary operator (denoted as \textit{div}). Finally, our grammar considers some constants that are commonly found in kernel functions as terminals. Although the White Noise kernel could be included as a terminal in the grammar, it is added to all the kernels generated during the search in order to model the noise (See Section~\ref{ssec:expsetup}).

\begin{table}[!ht]
  \centering
  \setlength\tabcolsep{4.5pt}
  \begin{tabular}{|c|c|c|c|c|}
    \hline
    & Symbol & Input & Expression & Output \\
    \hline
    \hline
    \parbox[t]{1.5mm}{\multirow{5}{*}{\rotatebox[origin=c]{90}{Args.}}}& x & & & $(\mathbf{x}, \mathbf{x}')$\\
    & $hp_0$ & & & $\theta$\\
    & $hp_1$ & & & $\theta$\\
    & ... & & & ...\\
    & $hp_t$ & & & $\theta$\\
    \hline
    \parbox[t]{1.5mm}{\multirow{7}{*}{\rotatebox[origin=c]{90}{Not Nestable}}}& euc & $(\mathbf{x}, \mathbf{x}')$ & \begin{tabular}[x]{@{}c@{}} $\mathbf{\hat{x}} = \mathbf{x}$ \\
    $\mathbf{\hat{x}}' = \mathbf{x}'$ \end{tabular} & $(\mathbf{\hat{x}},\mathbf{\hat{x}}')$ \\
    & spectral & $(\mathbf{x}, \mathbf{x}'), \theta_{p}$ & \begin{tabular}[x]{@{}c@{}}$ \mathbf{\hat{x}} = \begin{bmatrix}sin(\theta_{p} * \mathbf{x}) \\ cos(\theta_{p} * \mathbf{x}) \end{bmatrix}^T$ \\  $\mathbf{\hat{x}}' = \begin{bmatrix}sin(\theta_{p} * \mathbf{x}') \\ cos(\theta_{p} * \mathbf{x}') \end{bmatrix}^T$ \end{tabular} & $(\mathbf{\hat{x}},\mathbf{\hat{x}}')$ \\
    & sq\_dist & $(\mathbf{\hat{x}},\mathbf{\hat{x}}'), \theta_{l}$ & $z = \frac{1}{\theta_{l}^2} \left\lVert\mathbf{\hat{x}}-\mathbf{\hat{x}}'\right\rVert^2 $ & z\\
    & dot\_prod & $(\mathbf{\hat{x}},\mathbf{\hat{x}}'), \theta_{s}, \theta_{l}$  & $z = \frac{1}{\theta_{l}}\left(\mathbf{\hat{x}} - \theta_{s}\right) . \left(\mathbf{\hat{x}}' - \theta_{s}\right)^T$ & z\\
    & hp & $\theta$ & $z=\theta$ & z \\
    \hline
    \parbox[t]{1.5mm}{\multirow{7}{*}{\rotatebox[origin=c]{90}{Nestable}}}& power & $x, \theta$ & $z = w^{\theta}$ & z \\
    & add & $w, v$ & $z = w + v$ & z \\
    & multiply & $w, v$ & $z = w \times v$ & z \\
    & div & $w$ & $z=w^{-1}$ & z \\
    & exp & $w$ & $z=e^{w}$ & z \\
    & sqrt & $w$ & $z=\sqrt{w}$ & z \\
    & square & $w$ & $z=w^2$ & z \\
    \hline
    \parbox[t]{1.5mm}{\multirow{7}{*}{\rotatebox[origin=c]{90}{Terminals}}} & $-1$ & & $z=-1$ & z\\
    & $-0.5$ & & $z=-0.5$ & z\\
    & $0.5$ & & $z=0.5$ & z\\
    & $1$ & & $z=1$ & z\\
    & $2$ & & $z=2$ & z\\
    & $3$ & & $z=3$ & z\\
    & $5$ & & $z=5$ & z\\
    \hline
  \end{tabular} 
  \vspace{0.05cm}  
  \caption{Set of expressions used by the grammar. Column "Symbol" refers to the symbol given to each expression in the grammar. In column "Expression" the assignments ($=$) of the mathematical expressions to the outputs are shown. Finally, Columns "Input" and "Output" indicate the input and output types of each expression respectively. In these columns, $(\mathbf{x}, \mathbf{x})'$ represents the \textit{input vector} type, while $(\mathbf{\hat{x}}, \mathbf{\hat{x}}')$ shows the \textit{transformed input vector} type. The \textit{hyperparameter} type is denoted as $\theta$ and the $z, v, w$ variables belong to the \textit{cov} type.}
  \label{tab:grammar}
\end{table}



\section{Evolving kernel functions based on the new grammar}
\label{sec:kernelsearch}


Once the grammar has been introduced, we present our GenProg approach for GP kernel search, EvoCov. This algorithm takes into account two challenges related to this problem: The cost of evaluating the fitness function (mainly due to hyperparameter optimization) and the fact that many of the kernels generated during the search are not PSD.


Our GenProg approach is shown in Algorithm~\ref{alg:evogp}. First, an initial population of $N$ kernels is generated. In order to do so, each individual is created at random, limited by a maximum ($d_{max}$) and a minimum ($d_{min}$) depth. Each generation, the whole population is evaluated. Next, the relative improvement is measured. If the relative improvement in the current population is greater than a threshold $\beta$, a new population is generated through selection and variation. After selecting the $\mu$ best individuals, the algorithm randomly chooses the variation method between a mutation or a crossover operator (with probability $p_{m}$ and $p_{cx}$ respectively, where $p_{cx}=1-p_{m}$) to generate an offspring population of $N - \mu$ new individuals. The next population is made up of the selected individuals and the offspring population\footnote{Note that, some individuals may be evaluated several times as is explained in Section~\ref{sssec:hpinhe}.}. When the relative improvement is lower or equal to the threshold, the best individual is saved and the current population is replaced by a randomly generated one. This procedure is repeated for $G$ generations, until the last population is evaluated and the best individual found during the whole process is returned.

\begin{algorithm}
  \caption{EvoCov algorithm}
  \label{alg:evogp}
  \begin{algorithmic}[1]
    \Procedure{EvoCov}{$N$, $G$, $\mu$, $p_{m}$, $p_{cx}$, $\beta$, $d_{min}$, $d_{max}$}
      \State $best = \emptyset$
      \State $pop =$ \Call{GenRandPop}{$N$, $d_{min}$, $d_{max}$}
      \State $bestfitness_0 = \infty$
      \State $i = 1$
      \While{$i < G$}
        \State \Call{Evaluate}{$pop$}
        \State $bestfitness_i = \Call{BestFitness}{$pop$}$
        \State $relimprov = \frac{bestfitness_{i-1} - bestfitness_i}{|bestfitness_i|}$
        \If{$\beta < relimprov$} 
          \State $sel =$ \Call{Select}{$pop$, $\mu$}
          \State $offspring =$ \Call{Variate}{$sel$, $N - \mu$, $p_{m}$, $p_{cx}$}
          \State $pop = offspring \cup sel$
          \State $bestfitness_{i-1} = bestfitness_i$
        \Else \Comment{Restart procedure}
      	  \State $best =$ \Call{Select}{$pop \ \cup \{best\}$, $1$} 
          \State $pop =$ \Call{GenRandPop}{$N$, $d_{min}$, $d_{max}$}
          \State $bestfitness_{i-1} = \infty$
        \EndIf
        \State $i = i + 1$
      \EndWhile
      \State \Call{Evaluate}{$pop$}
      \State $best =$ \Call{Select}{$pop \ \cup \{best\}$, $1$} 
      \State \Return $best$
    \EndProcedure
  \end{algorithmic}
\end{algorithm}

In this section we describe each of the methods used by Algorithm~\ref{alg:evogp}. First, we address the issue of randomly generating new kernels for the initial population. The distinguished characteristic of our proposal for generating the random kernels is that it does not take into account any kernel proposed in the literature, while guaranteeing a minimum depth and a maximum depth. Then, we provide the variation operators conceived to generate kernels that are likely to inherit useful properties from the selected ones. A method to control the depth of the trees created by the variation operators is also introduced. Next, we explain how the GenProg kernels are evaluated. Finally, for comparison proposes, we include two simpler methods: Random Search and Go With The First.


\subsection{Initial population}
\label{ssec:initpop}

We generate kernel functions at random and discard the non-PSD ones until the desired population size is reached.



\subsubsection{Random generation}
\label{sssec:randgen}

We propose a strongly-typed grow method to randomly generate kernel expression trees, based on the work done in \cite{koza_genetic_1992}. This is achieved by a recursive process where, at each step, a random terminal or operator is added.

When randomly generating solutions, some of the solutions may be small and trivial, and some other may be too complex. Thus, we propose a method to control the depth of the generated trees by setting a minimum ($d_{min}$) and a maximum depth ($d_{max}$). As can be seen in Table~\ref{tab:grammar}, some of the expressions have at least one input that matches the output type. These expressions guarantee that, once selected, the iterative procedure can continue growing this branch, i.e., these expressions are nestable.

As can be seen in Algorithm~\ref{alg:rand_gen}, during the recursive process, we select a random expression depending on the current type. If the minimum depth has not been reached, only the operators that can be nested are used. Then, until the maximum depth is reached, any operator can be selected. Finally, when the maximum depth is reached, only the terminals, arguments and the operators that can not be nested are used, limiting the depth of the tree.

\begin{algorithm}
  \caption{Random Generation of expression trees}
  \label{alg:rand_gen}
  \begin{algorithmic}[1]
    \Procedure{TypedGrow}{$d_{min}$, $d_{max}$, $type$}
      \State $terms =$ \Call{GetTerminalsAndArgs}{$type$}
      \State $notnests =$ \Call{GetNotNestable}{$type$}
      \State $nests =$ \Call{GetNestable}{$type$}
      \State $candexprs = \emptyset$
      \If{$d_{min} <= 3$}
      	\State $candexprs = candexprs \cup notnests$
        \If{$d_{min} <= 1$}
          \State $candexprs = candexprs \cup terms$
        \EndIf
      \EndIf
      \If{$4 <= d_{max}$}
      	\State $candexprs = candexprs \cup nests$
      \EndIf
      \If{$candexprs$ \textbf{is} $\emptyset$} \Comment{No candidate expressions}
      	\State $candexprs = terms \cup notnests \cup nests$
      \EndIf
      \State $expr =$ \Call{RandomChoice}{$candexprs$}
      \If{\Call{nInput}{$expr$} $== 0$}
      	\State \Return $expr$
      \EndIf
      \For{$input$ \textbf{in} $expr$}
        \State $inputtype =$ \Call{Type}{$input$}
        \State $subexpr =$ \Call{TypedGrow}{$d_{min}-1$, $d_{max}-1$, $inputtype$}
        \State \Call{Append}{$expr$, $subexpr$} 
      \EndFor
      \State \Return $expr$
    \EndProcedure
  \end{algorithmic}
\end{algorithm}

\subsubsection{Checking positive semi-definiteness}
\label{sssec:check}

Non-PSD kernels should be excluded from our search space as they are not valid for GP. Although it is computationally unfeasible to guarantee that every single kernel is PSD, we follow an efficient method to identify most of the non-PSD kernels, similar to the work carried out in the SVM literature \cite{koch_tuning_2012, diosan_evolving_2007, howley_evolutionary_2006}.

As mentioned in Section~\ref{ssec:kernel}, any covariance matrix generated by a PSD kernel has to be symmetric and also PSD. To identify non-PSD kernels, we generate $w$ random uniformly distributed data sets $X=(\mathbf{x}_1,\mathbf{x}_2,...,\mathbf{x}_n)$ (where $\mathbf{x}_i \in \mathbb{R}^d$, $i \in \{ 1,...,n \}$  and $n\in\mathbb{N}$) and check the covariance matrix $C$ produced by the kernel for each data set. If any covariance matrix matches the following cases, the kernel is rapidly discarded:

\begin{itemize}
  \item $C \neq C^T$: As previously mentioned, the covariance matrix given by a PSD kernel should be symmetric.
  \item Any $c_{ii}$ is negative: It has been proved \cite{zhang_positive_2011} that, if any of the elements in the main diagonal are negative, the covariance matrix is not PSD.
  \item Any of the eigenvalues of $C$ is negative: Similarly, all the eigenvalues of the covariance matrix should be non-negative.
\end{itemize}

Not matching these cases is necessary but not sufficient for a kernel to be PSD. If after meeting this condition, during the evaluation step, we find out that the kernel is not PSD, the fitness value of the kernel is penalized. However, this validity check is severe enough to avoid most of the false positives. Among the kernels that were generated and validated during the experiments conducted in this paper, only $0.67\%$ were not evaluable.

\subsection{Variation operators for kernel generation}
\label{ssec:variation}

Our kernel search method is based on perturbation or variation methods that modify previous solutions to obtain new ones. We use two variation operators, which are randomly selected every VARIATE function call in Algorithm~\ref{alg:evogp}: A crossover operator, which combines two kernel functions to generate a new one that keeps some of the features of its parents, and a mutation operator, which introduces slight modifications to the original kernel to obtain a new individual. We also explain how we control the depth of the trees generated by these variation methods.

\subsubsection{Crossover}
\label{sssec:Crossover}

A purely random crossover operator hardly ever produces PSD kernels. Since kernel function evaluation is a computationally costly process, we would like to avoid non-PSD kernels. As explained in Section \ref{ssec:kernelcomp}, the product or the sum of two PSD kernels is also PSD. Hence, a crossover method could just combine two PSD kernels with any of these operators to generate a new PSD kernel. However, this procedure rapidly increases the depth of the expression trees. 

Therefore, we propose a crossover operator that randomly selects a subtree from each kernel and combines them with the sum or the product operator. As this method does not guarantee that the resulting kernel is PSD, this operation must be repeated if a non-PSD kernel is found. Nevertheless, the method increases the chance of obtaining a PSD kernel, since if both of the subtrees are PSD, the result is guaranteed to be also PSD.



\subsubsection{Mutation}
\label{sssec:mutation}

The mutation operator works by randomly selecting one of the following methods in a type-safe manner:
\begin{description}
  \item [Insert:] Inserts an elementary mathematical expression (see Table~\ref{tab:grammar}) at a random position in the tree, as long as their output types agree. The subtree at the chosen position is used as the input of the created expression. If more inputs are required by the new primitive, new terminals are chosen at random.
  \item [Shrink:] This operator shrinks the expression tree by randomly choosing a branch and replacing it with one of the branch inputs (also randomly chosen) of the same type.
  \item [Uniform:] Randomly selects a point in the expression tree and replaces a subtree at that point by the subtree generated using our random generation method (Described in detail in Section~\ref{sssec:randgen}). Note that the output type of the random generated subtree must match the output type of the replaced one.
  \item [Node Replacement:] Replaces a randomly chosen operator from the kernel expression by an operator with the same number of inputs and types, also randomly chosen.
\end{description}

As these methods do not guarantee that the generated kernels are PSD, mutations are repeated if a non-PSD kernel is detected (see Section~\ref{sssec:check}).


\subsubsection{Bloat control}
\label{sssec:bloat}

None of the variation methods limit the depth of the kernel expression. Depending on the operators, the depth of the expression trees may increase without any limit during the search, making the resulting kernel functions complex and useless for practical applications. This is a well known problem in GenProg literature, known as bloating \cite{koza_genetic_1992}. In our work, when the depth of a kernel expression becomes larger than $o_{max}$, we discard the expression. In this case, the mutation or the crossover method is repeated until a kernel with the desired depth is obtained, or a limit of $\rho_{max}$ trials is reached. If this number of trials is exceeded, one of the parent kernels is returned unchanged.





\subsection{Evaluation}
\label{ssec:evaluation}

\begin{figure}[!ht]
    \centering
    \includegraphics[width=0.65\textwidth]{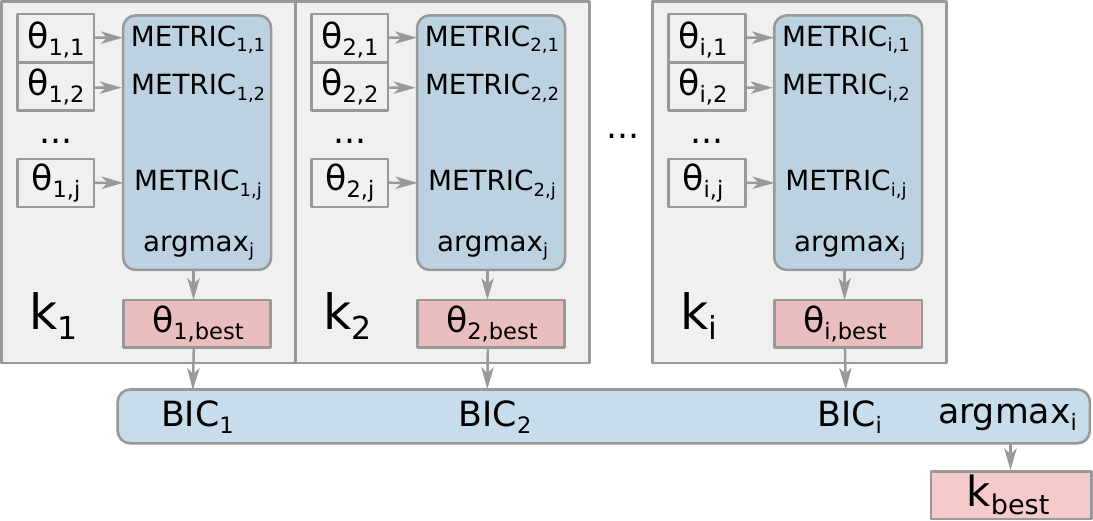}
    \caption{Two nested search procedures: The selection of the best hyperparameters for each kernel is made according to a given metric, such as LML or LOOCV, and the selection of the best kernel according to the BIC.}
    \label{fig:kernel_search}  
\end{figure}

In our approach, in contrast to other GenProg applications, the solutions do not encode all the necessary information to be evaluated. In order to evaluate each kernel, we need to set the value of the hyperparameters. Thus, the fitness of the solutions depends on the results of the hyperparameter optimization. Both search procedures, the selection of the best hyperparameters for each kernel and the selection of the best kernel given these hyperparameters, are illustrated in Figure~\ref{fig:kernel_search}.

As in \cite{duvenaud_structure_2013}, we use the Bayesian Information Criterion (BIC) \cite{schwarz_estimating_1978} as a quality metric for each kernel. BIC is a metric for model selection which adds a regularization term to the LML to penalize the complexity of the kernels. This metric serves as the fitness function of our GenProg algorithm and it can be expressed as follows:

\begin{equation} \label{eq:bic}
\begin{split}
BIC(k_i) &= -2 \; log \; p \left( \mathbf{f}| X, k_i, \boldsymbol\theta_{i, best} \right) + q \; log \; n
\end{split}
\end{equation}
where $q$ is the number of hyperparameters of the kernel and $n$ is the number of data points in $X$. $\boldsymbol\theta_{i, best}$ is the best hyperparameter set for the kernel $k_i$ according to a given metric.

Before computing the BIC associated to a given kernel, the hyperparameters have to be optimized. As we have seen in Section~\ref{ssec:modelsel}, several metrics (LML, LOOCV, ...) can be used to measure the quality of each hyperparameter set. Thus, we find the best hyperparameter set for kernel $i$ as follows:
\begin{equation} \label{eq:pri_thetaopt}
\boldsymbol\theta_{i, best} = argmax_{j} \; \textmd{METRIC} \left( \mathbf{f}, X, k_i, \boldsymbol\theta_{i, j} \right)
\end{equation}

\subsubsection{Hyperparameter Optimization Algorithm}
\label{sssec:hpopt}

The hyperparameters are optimized by means of \textit{Powell}'s local search algorithm \cite{powell_efficient_1964}. As this algorithm is not bounded, the search space has to be constrained by penalizing non-feasible hyperparameter sets. On the other hand, as the function to optimize might be multi-modal, a multi-start approach was used, performing a restart every time the stopping criteria of the \textit{Powell}'s algorithm are met, and getting the best overall result. Note that, as a result of the inclusion of the randomized restarts, the hyperparameters found for a certain kernel in two independent evaluations may not be the same. In fact, this implies that the fitness function optimized by the GenProg algorithm, i.e., BIC, is stochastic.

\subsubsection{Random Restarts and Hyperparameter inheritance}
\label{sssec:hpinhe}
The initial solutions for the restarts of the hyperparameter optimization algorithm are sampled from two different distributions depending on the origin of the kernel. In the randomly generated kernels, the initial hyperparameters for these restarts are sampled from a uniform distribution within the search bounds. On the other hand, if a kernel is generated through any of the variation methods, we take advantage of the information gathered in previous hyperparameter optimization procedures by adapting the inheritance technique described in \cite{duvenaud_structure_2013} and \cite{lloyd_automatic_2014} to the particularities of GenProg. Instead of restarting the multi-start optimization from a uniform distribution, each restart is sampled from a Gaussian distribution centered on the hyperparameter values of the parent individuals and with a pre-defined variance ($\sigma_{\theta}$). This inheritance method is particularly useful when the variation performs few changes to the expression tree.

Note that, in Algorithm~\ref{alg:evogp}, the selected individuals are kept for the next population and the whole population is evaluated each generation. Thus, some individuals may be evaluated several times during the search. This procedure, along with the hyperparameter inheritance, allows the selected individuals to keep optimizing their hyperparameters across generations, and compete fairly with the individuals in the offspring population, which inherit the hyperparameters.

\subsection{Selection}
\label{ssec:selection}

We perform a search in the kernel function space to find the kernel that maximizes the BIC. Thus, the selection operator shown in Algorithm~\ref{alg:evogp} selects the $\mu$ best kernels according to the BIC metric by applying truncation selection.

\subsection{Alternative search methods}
\label{ssec:altmethods}

In order to verify that every component of the proposed GenProg algorithm is providing a benefit to the kernel search, we introduce two algorithms to be used as a baseline in the experiments. First, we describe a Random Search algorithm to test the contribution of the components of EvoCov with the exception of the random generation method. Then, we propose a Go With The First algorithm, which does not depend on the crossover operator, in order to measure the gain produced by this operator.

\subsubsection{Random Search}
\label{sssec:random_search}

This Random Search method generates a random population by iteratively following the random generation method described in Section~\ref{sssec:randgen} until the desired population size is achieved ($N$). Next, it chooses the best solution according to the selection criterion described in Section~\ref{ssec:selection}.


\subsubsection{Go With The First}
\label{sssec:go_with_first}


This algorithm generates an initial population of size $G$ and, for each individual, a hill-climbing procedure is applied for $N$ evaluations. This procedure is carried out by generating a random mutation, and keeping the best solution between the original and the mutated one. Once all the individuals in the population have been optimized following this procedure, the worst one according to the selection criteria is discarded. This whole process is repeated until only one kernel is left.


\section{Related work}
\label{sec:relwork}

In this section, we review the work carried out in the literature related to this paper.  First, we discuss works that design ad hoc kernels based on expert knowledge. Subsequently, we review the works which propose an automatic design of kernels.

In ad hoc kernel approaches \cite{rasmussen_gaussian_2006,preotiuc-pietro_temporal_2013}, the authors assume that the choice of the kernel function is clear from a priori knowledge about the problem. Then, the hyperparameters are optimized to adjust each kernel. In \cite{rasmussen_gaussian_2006}, an ad hoc kernel is introduced to fit the Mauna Loa Atmospheric $CO_2$ time-series, which is a well-known problem in the GP literature due to its several periodic patterns. On the other hand, in \cite{klenske_nonparametric_2013}, the authors propose a product of a Squared Exponential and a periodic kernel to construct a control signal. Similarly, the authors of \cite{preotiuc-pietro_temporal_2013} designed an ad hoc kernel to predict the number of occurrences of certain hashtags in Twitter, given the past records. Finally, the authors of \cite{wilson_gaussian_2013} took advantage of the Bochner Theorem \cite{bochner_lectures_1959} to design kernels that were able to model the periodical patterns of time-series. Regarding the hyperparameter optimization, Deep Learning methods have also been applied to pre-train and fine-tune the hyperparameters of the covariance functions \cite{hinton_using_2008}.

Regarding the automatic design of kernels, many works have followed the kernel composition approach by using the properties shown in Section~\ref{ssec:kernelcomp} to generate new kernels \cite{kronberger_evolution_2013, duvenaud_structure_2013}. Authors of \cite{kronberger_evolution_2013} propose a GenProg method for compositional kernel search, using the well-known kernels shown in Table~\ref{tab:kern} as building blocks. They also consider the sum, product and scale as primitives, along with a dimension mask. The hyperparameters where not included in the grammar, as only the hyperparameters present in the well-known kernels are considered. The experimentation of this work was limited to the Mauna Loa Atmospheric $CO_2$ time-series and some synthetic 2-dimensional data sets. In addition to GenProg, other search methods have been proposed to search for kernel composition structures in GPs, such as the greedy search procedure proposed in \cite{duvenaud_structure_2013}. In that work, the best kernel function in terms of BIC is searched in the space of possible compositions (sums and products) of simpler kernels. In \cite{lloyd_automatic_2014}, the authors  improve the previous approach by adding change-point and change-window kernels. Finally, in \cite{malkomes_bayesian_2016}, Bayesian Optimization was used to search in the model space.

The idea of using elementary mathematical expressions as building blocks of kernel functions has been applied to other fields, such as Support Vector Machines (SVMs) \cite{diosan_evolving_2007} and Relevance Vector Machines (RVMs) \cite{bing_gp-based_2010}. Some of these approaches \cite{bing_gp-based_2010,howley_genetic_2005,gagne_genetic_2006} do not guarantee that kernels are PSD. In SVM and RVM, although kernels theoretically have to be PSD in order to do the kernel-trick, in practice, many kernels can be used even if they are not of this kind. Some other approaches, such as \cite{diosan_improving_2012} and \cite{sullivan_evolving_2007}, guarantee that the kernels are PSD by means of the kernel composition properties shown in Section~\ref{ssec:kernelcomp}, similar to the compositional kernel search methods in the GP literature. Finally, in \cite{koch_tuning_2012}, \cite{diosan_evolving_2007} and \cite{howley_evolutionary_2006} the non-PSD kernels are penalized or discarded as in our approach. The authors of \cite{diosan_evolving_2007} and \cite{howley_evolutionary_2006} propose a method to penalize (giving the worst possible fitness) the non-PSD kernels in evaluation time. On the other hand, in \cite{koch_tuning_2012}, if during the random generation a non-PSD kernel is found, it is discarded and the creation is retried. However, as stated by the authors, their approach was not able to improve the results of the standard kernels in SVM. The above mentioned approaches deal with hyperparameters by means of optimizing small grids or adding random constants to the GenProg grammar. To the best of our knowledge, more complex techniques such as the hyperparameter inheritance have not been applied in this context.
  

 

\section{Experiments}
\label{sec:experiments}

In this section, we describe the experiments we carried out to analyze the performance of our proposal. We solve extrapolation problems from real-world time-series and compare our proposal to the main methods discussed in Section~\ref{sec:relwork} (compositional kernel methods and ad hoc kernel approaches) in such tasks. The goal of our experiments is three-fold:

\begin{itemize}
  \item To compare EvoCov to state-of-the-art methods that rely on kernel composition. 
  \item To compare our proposal to the ad hoc kernels proposed in the literature.
  \item To study the influence of the metric used to optimize the hyperparameters in time-series extrapolation problems.
\end{itemize}

First, an introduction to the time-series extrapolation problem is given, before describing the experimental setup. Then, three experiments are shown, one for each objective of the experimentation.





\subsection{Time-series extrapolation problems}
\label{ssec:tsextrap}

The objective in time-series extrapolation is to predict future time-stamp values given some previous data. While properties like the smoothness of the data have been extensively studied in GP literature for interpolation problems, other properties required in extrapolation have not been studied to the same extent, such as periodicities and trends.

Real-world time-series problems have been considered for the evaluation of our methods, as they are more realistic than the synthetic environments. The selected problems are characterized by a limited amount of usually noisy data with strong variations between training and test sets.



In Table~\ref{tab:timeseries}, the real-world time-series used in the first two experiments are described\footnote{The time-series data can be found at \url{https://datamarket.com/data/list/?q=provider:tsdl}}. Following the work done in \cite{lloyd_automatic_2014}, we have trained all the algorithms on the first 90\% of the data, predicted the remaining 10\%, and then computed the root mean squared error (RMSE) for that 10\%. 

\begin{table}[!ht]
  \centering
  \begin{tabular}{|l|rl|}
    \hline
    Name &   Size  & Properties \\
    \hline
    \hline
      Airline    & 144 & P, AT \\
      Solar      & 402 & P, C   \\
      Mauna Loa Atmospheric $CO_2$  & 545 & P, AT     \\
      Beveridge Wheat Price Index & 370 & P, C \\
      Daily minimum temperatures in Melbourne  & 1000 & P N\\
      Internet traffic data (bits) & 1000 & P+ \\
      Monthly average daily calls & 180 & N, C \\
      Monthly critical radio frequencies & 240 & P+  \\
      Monthly production of gas in Australia & 476 & P, AT\\
      Monthly prod. of sulphuric acid in Australia & 462 &  N \\
      Monthly U.S. male unemployment & 408 &  P, AT \\
      Number of daily births in Quebec & 1000 &  N \\
      Real daily wages in England ($\pounds$) & 735 & EXP \\
    \hline
  \end{tabular}
  \vspace{0.15cm}
  \caption{Description of the time-series used in the experiments. The visually identifiable periodic patterns are described with the letter P, and if many periods are present, P+ is shown. The ascendant trends are represented by AT and exponential growths by EXP. N denotes the presence of noise, while C indicates a trend change at some point in the time-series.}
  \label{tab:timeseries}
\end{table}

\subsection{Experimental Setup}
\label{ssec:expsetup}

Our algorithms were coded in Python, based on the EA software \texttt{DEAP}\footnote{\url{https://deap.readthedocs.io}} \cite{fortin_deap:_2012}. For the GP regression, a noisy approach was used, adding a white noise kernel to all the generated kernels, including a noise hyperparameter. For the random generation method, $d_{min}=5$ and $d_{max}=15$ were used to limit the size of each expression tree. In order to discard the non-PSD kernels, the positive semi-definiteness conditions described in Section~\ref{sssec:check} were checked in $w=20$ random data sets. On the other hand, to avoid bloating, a maximum depth of $o_{max}=40$ was allowed, and the number of attempts was limited to $\rho_{max}=250$ in each variation operator.

In the Random Search algorithm, $N=20000$ was set to generate the initial population. Similarly, in the Go With The First algorithm, $G=13$ generations and $N=200$ local search evaluations were used in order to have a comparable evaluation budget. Finally, in the EvoCov approach, $N=141$, $G=141$, $\mu=14$, $p_{m}=0.4$, $p_{cx}=0.6$ and $\beta=1\mathrm{e}{-5}$ were set. Due to the stochastic nature of our algorithms, each kernel search process was repeated 10 times in all the experiments. 

Regarding the hyperparameter optimization, every restart of the \textit{Powell's} optimization algorithm, a Gaussian noise with $\sigma_{\theta}=0.1$ was added to the inherited hyperparameters. Since the computational time required to evaluate the hyperparameters increases quadratically with the size of the time-series, in order to keep the computational cost of optimizing these hyperparameters similar in all the problems, we decided to adjust the number of evaluations allowed in the hyperparameter optimization depending on the length of the time-series. Thus, we allow $max\_fun\_call = ref\_fun\_call * \frac{350^2}{current\_ts\_len^2}$ evaluations, where $current\_ts\_len$ is the length of the current time-series and $ref\_fun\_call$ is a parameter of the algorithm. In the experiment shown in Section~\ref{ssec:hpoptmetrics}, we allow $ref\_fun\_call=5000$ evaluations, and in the rest of the experiments $ref\_fun\_call=300$.

\subsection{Metric comparison for hyperparameter optimization}
\label{ssec:hpoptmetrics}

In GP literature, hyperparameter optimization is considered a crucial task. Most of the works carried out in this field rely on the LML for hyperparameter optimization \cite{kronberger_evolution_2013,duvenaud_structure_2013}. However, it has been reported that the LML may lead to suboptimal results under certain conditions, where LOOCV could be more robust\cite{bachoc_cross_2013}. Regarding the kernel optimization, having a consistent method to optimize the hyperparameters helps to obtain more reliable evaluations of the individuals. Hence, we decided to perform a preliminary experiment to test if other alternative metrics to LML and LOOCV can improve the results in time-series extrapolation problems.

Apart from the well-known LML and LOOCV, we tested other metrics specifically designed to optimize the hyperparameters in extrapolation problems. While the LML measures the probability of the training data given the prior GP, the goal in extrapolation is to increase the probability of the test data given the posterior GP. Thus, along with the prior LML, we also measure the posterior LML. By splitting the train set into some train-train and train-test sets, we get the posterior GP given the train-train set, and measure the probability of the train-test set given this model, i.e., the posterior LML. On the other hand, the LOOCV measures the ability of the GP to interpolate. To create a version of the LOOCV metric for extrapolation, we compute the sum of the posterior probabilities of each of the data points in the train-test set, and call it Sum of Posterior Likelihoods (SoPL). Finally, having the train-train and train-test sets, the RMSE in the train-test set is also used as a measure given the train-train set.

To compare the performance of these metrics, we considered the best kernels found in \cite{duvenaud_structure_2013} for all the time-series described in Section~\ref{ssec:tsextrap}. For each metric and time-series, we carried out an optimization process with the Powell's algorithm, in order to find which one leads to the best results in terms of the RMSE in the test set. Each optimization process starts from a random hyperparameter set and stops when 5000 samples have been taken. Due to the randomness of the process, 10 trials were carried out.

In Table~\ref{tab:hpopt_results}, the average results are shown for the thirteen time-series. The SoPL outperforms the rest of the metrics in five of the problems, while LML gets the best overall results in four time-series and RMSE is the best choice in three. Those three metrics are superior to posterior LML and LOOCV, as the former is only able to obtain the best results in the \textit{Daily Minimum Temperatures in Melbourne} time-series, and the latter is not able to beat other metrics in any of the problems. As expected, in these extrapolation problems LOOCV is the metric with the worst performance.

\begin{table*}[!ht]
  \centering
  \begin{tabular}{|l|rrrrr|}
    \hline
    &   -LML      & -LOOCV                              & -Post. LML   & -SoPL & RMSE\\
    \hline
    \hline
    Airline                         & \textbf{37.00}    & 230.27     & 157.96     & 57.10    & 97.34    \\
    Solar                           & 269.12   & 539.64     & 925.25     & 279.24   & \textbf{132.04}   \\
    Mauna Loa Atmospheric $CO_2$                       & 4.40     & 37.61      & 3.94       & \textbf{2.34}     & 3.19     \\
    Beveridge Wheat Price Index   & 54.13    & 99.64      & 67.94      & \textbf{52.58}    & 266.37   \\
    Daily min. temps. in Melbourne   & 4.92     & 6.63       & \textbf{4.62}       & 5.62     & 4.67     \\
    Internet traffic data (bits) & 49352.14 & 4994073.33 & 38259.26   & \textbf{23756.89} & 25970.23 \\
    Monthly average daily calls   & 212.22   & 43078.20   & 1460.47    & 844.32   & \textbf{55.08}    \\
    Monthly critical radio frequencies   & 2.14     & 2.01       & 1.27       & \textbf{0.74}     & 1.63     \\
    Monthly prod. of gas in Aus.   & \textbf{13791.03} & 179944.80  & 26403.55   & 18066.63 & 50771.39 \\
    Monthly prod. of sul. acid in Aus.   & \textbf{39.58}    & 1979.48    & 56.56      & 53.42    & 67.89    \\
    Monthly U.S. male unemployment   & \textbf{142.30}   & 4436.13    & 265.76     & 192.34   & 219.65   \\
    Number of daily births in Quebec   & 44.78    & 16203.27   & 49.84      & \textbf{44.54}    & 46.40    \\
    Real daily wages in England ($\pounds$)   & 23.26    & 40.97      & 15.16      & 15.38    & \textbf{13.61}    \\
    \hline
  \end{tabular}
  \vspace{0.15cm}
  \caption{Hyperparameter optimization metrics compared across different time-series. The average RMSE in the test set is shown for each metric. Smaller is better in all cases, as the negative LML, LOOCV, posterior LML and SoPL are shown. The best results for each time-series are shown in bold. }
  \label{tab:hpopt_results}
\end{table*}

\begin{figure}[!ht]
    \centering  
    \includegraphics[trim={2.10cm 2.75cm 1.75cm 5cm},clip,width=0.4\textwidth]{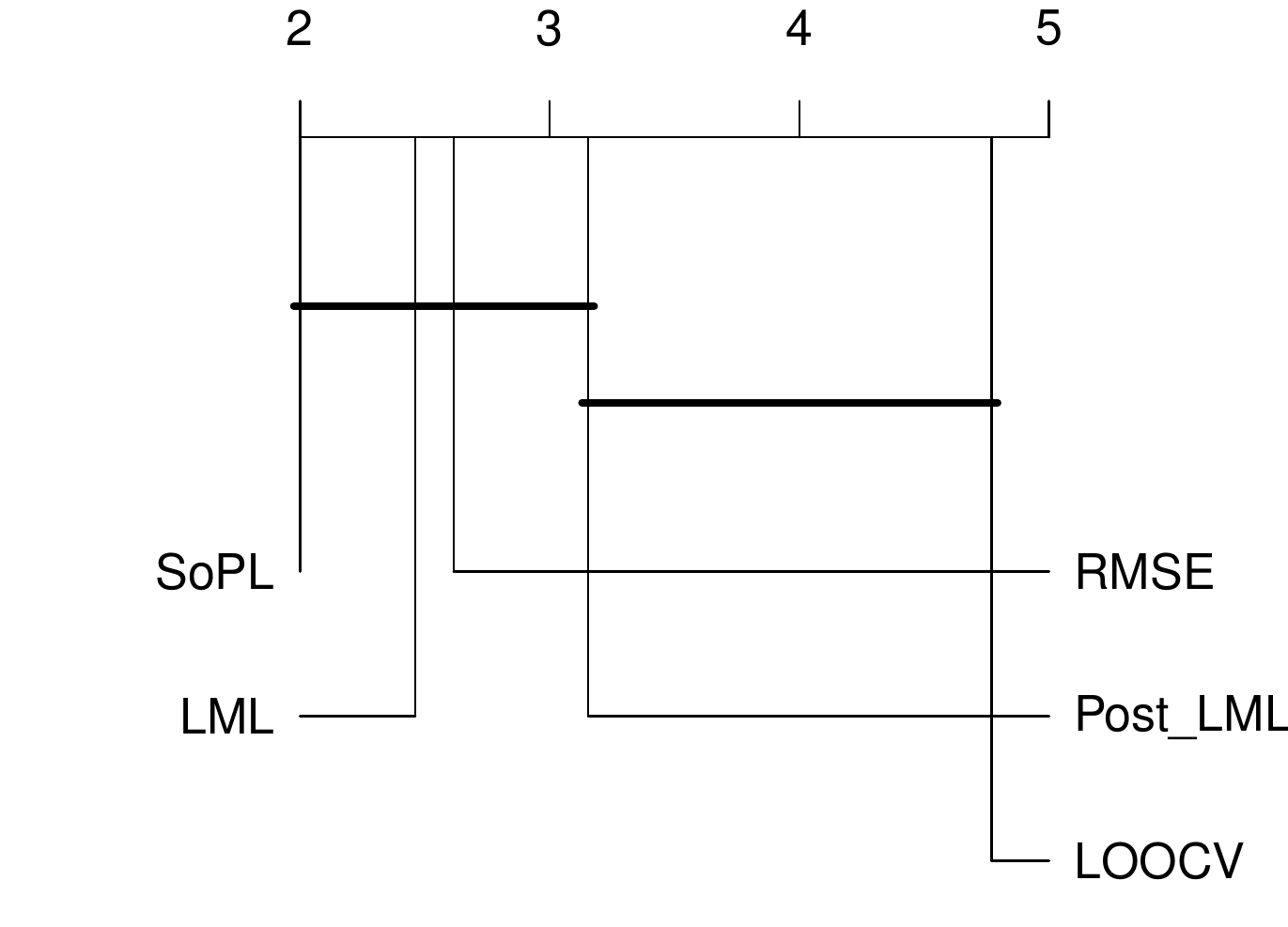}
    \caption{Critical differences diagram. The metrics are ordered following the results in their ranking. The metrics with no significant differences between them are matched with a straight line.}
    \label{fig:critical_diff_metrics}
\end{figure}

Statistical tests were used to determine if there is a metric that is more robust than the others in time-series extrapolation problems\footnote{The tests were carried out using the SCMAMP R package \cite{calvo_scmamp:_2016}}. First, we averaged all the RMSE results for each metric and time-series, and applied the Friedman's test \cite{friedman_use_1937}. We found significant differences between all the metrics (p-value = $8.454\mathrm{e}{-5}$). Then, we applied a post-hoc test based on Friedman’s test, and adjusted its results with the Shaffer's correction \cite{shaffer_modified_2012}. In Figure~\ref{fig:critical_diff_metrics}, the critical differences between the metrics are shown. As can be seen, there are no significant differences between SoPL, LML, RMSE and posterior LML. Similarly, the results between LOOCV and posterior LML do not differ significantly.

Overall, it can be seen that there is no metric best suited for all the time-series, and the choice of the best metric depends on the problem.

As the differences between LML and SoPL are not significant, we used two variants of EvoCov for the following experiments, one using LML to optimize the hyperparameters (EvoCov LML) and the other using SoPL (EvoCov SoPL).

\subsection{Comparing our proposal to compositional kernel search approaches}
\label{ssec:vscks}

In kernel learning tasks, the compositional kernel search approaches hold the state-of-the-art results in the GP literature. In this section we describe the exploratory experiment we carried out initially, and the benchmark where we compare our algorithms to the compositional kernel search approaches.


\subsubsection{Initial Experiment}
\label{sssec:firstexp}

To begin with, we performed an initial experimentation to test whether better kernels can be found with our grammar, compared to the kernel composition approaches. Particularly, we would like to know:
\begin{enumerate}
  \item Whether it is possible to improve the composed kernels by means of manipulating elementary mathematical expressions, as we propose in this paper.
  \item Whether the tree-based representation of kernels, along with the variation operators, allow such an improvement.
\end{enumerate}

Given the best kernels found in \cite{duvenaud_structure_2013} for all the time-series, we generated 200 random mutations according to the method described in Section~\ref{sssec:mutation}. Next, we performed a hyperparameter optimization process using the LML metric for each mutation, departing from the best hyperparameter values found in the original work. Finally, we measured the RMSE in the test set for all these mutations against the RMSE provided in \cite{duvenaud_structure_2013}.

\begin{figure}[!ht]
    \centering   
    \includegraphics[trim={0cm 0.5cm 0cm 0.5cm},width=0.6\textwidth]{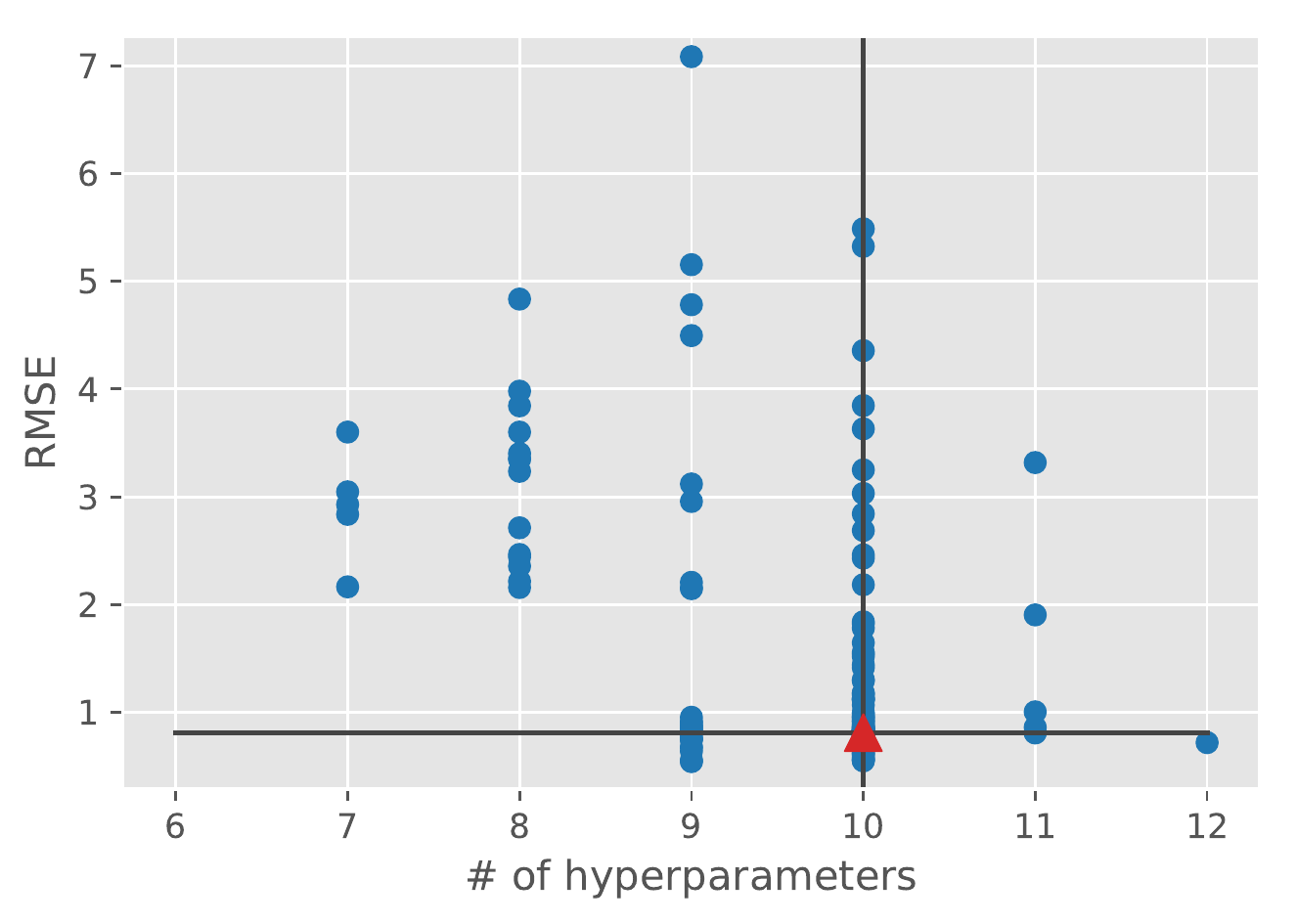}
    \caption{RMSE and number of hyperparameters of the 200 random mutations for  the Mauna Loa Atmospheric $CO_2$ time-series. The dark triangle represents the original kernel, and the straight lines represent its number of hyperparameters and RMSE.}
    \label{fig:pareto_in_neighbors}  
\end{figure}

 
As can be seen in Figure~\ref{fig:pareto_in_neighbors}, there are many mutated kernels that obtain a better RMSE than the original one for the Mauna Loa Atmospheric $CO_2$ time-series. Moreover, some of those kernels have fewer hyperparameters than the best kernel achieved in \cite{duvenaud_structure_2013} (9 instead of 10).
 
In Table~\ref{tab:intial}, similar results to the Mauna Loa Atmospheric $CO_2$ can be found for the rest of the time-series. Among the 200 randomly generated kernels in each of the time-series, there are some kernels that are better fitted according to the RMSE. Furthermore, in 12 out of 13 time-series there are mutated kernels that have better results in RMSE, with fewer hyperparameters.

\begin{table}[!ht]
  \centering
  \begin{tabular}{|l|rrr|}
    \hline
    Name &  Better fitted & Simpler & Both \\
    \hline
    \hline
    Airline                         & 0.38             & 0.43    & 0.13                 \\
    Solar                           & 0.55             & 0.32    & 0.21                 \\
    Mauna Loa Atmospheric $CO_2$                       & 0.18             & 0.34   & 0.05                 \\
    Beveridge Wheat Price Index   & 0.40            & 0.19   & 0.11                \\
    Daily min. temps. in Melbourne   & 0.23             & 0.27   & 0.00                    \\
    Internet traffic data (bits) & 0.24             & 0.38    & 0.07                 \\
    Monthly average daily calls   & 0.50              & 0.48    & 0.20                  \\
    Monthly critical radio frequencies   & 0.36            & 0.30   & 0.05                 \\
    Monthly prod. of gas in Aus.   & 0.71             & 0.41    & 0.30                  \\
    Monthly prod. of sul. acid in Aus.   & 0.56             & 0.43   & 0.23                 \\
    Monthly U.S. male unemployment   & 0.47             & 0.38    & 0.15                \\
    Number of daily births in Quebec   & 0.48             & 0.32    & 0.13                 \\
    Real daily wages in England ($\pounds$)   & 0.38            & 0.28   & 0.08   \\
    \hline
  \end{tabular}
  \vspace{0.15cm}
  \caption{Results of the initial experiment. In the first column, the ratio of mutated kernels that are better fitted, in terms of RMSE, than the best kernel achieved in \cite{duvenaud_structure_2013} is shown. The ratio of kernels that are simpler, according to the number of hyperparameters, is illustrated in the second column. In the last column, the ratio of kernels that are both better fitted and simpler can be found.}
  \label{tab:intial}
\end{table}

As we have shown in this exploratory experiment, we conclude that it is possible to improve the results obtained by compositional kernel search approaches by means of manipulating elementary mathematical expressions. We can also confirm that the mutation operator presented in this work allows such improvements.

\subsubsection{Benchmark}
\label{sssec:benchmark}

In view of results obtained in the previous section, we tested EvoCov, along with the proposed alternative search methods, in the benchmark presented in \cite{lloyd_automatic_2014}. To the best of our knowledge, this work provides the most extensive comparison in the literature in time-series extrapolation with GPs. In this benchmark the following algorithms can be found:

\begin{description}
  \item[Eureqa:] A Symbolic Regression engine that uses genetic algorithms to search in the space of the possible equations \cite{schmidt_eureqa_2013}. Although this approach may seem similar to our work, Eureqa learns the predictive function itself, while our approach provides a probabilistic prediction by means of a GP.
  \item[Linear Regression (LIN):] The basic linear regression is approximated by a GP with a Linear kernel. The hyperparameters are learned by the LML optimization. 
  \item[Squared Exponential (SE):] A GP with the Squared Exponential kernel shown in Table \ref{tab:kern} is used. The hyperparameters are also learned by optimizing the LML.
  \item[Bayesian variant of multiple kernel learning (MKL):] A weighted sum of base kernels is used to construct more complex ones \cite{bach_multiple_2004}. 
  \item[Change Point (CP) Modeling:] A GP based approach allowing changepoints in kernels, that is, a combination of kernels where the weight of each of the components depends on the inputs \cite{garnett_sequential_2010,saatci_gaussian_2010,fox_multiresolution_2012}.
  \item[Spectral Mixture Kernels (SK):] These kernels model the spectral density with a Gaussian mixture \cite{wilson_gaussian_2013}.
  \item[Trend-Cyclical-Irregular (TCI) Models:] The statistical model described in \cite{lind_basic_2006} is approximated by means of a GP and combining the periodic kernels with linear ones as covariance function.
  \item[GPSS:] The greedy GP kernel search method described in \cite{duvenaud_structure_2013} is used, as discussed in Section~\ref{sec:relwork}.
  \item[ABCD accuracy:] An improvement of GPSS, introduced in \cite{lloyd_automatic_2014}, which includes the ChangeWindow and ChangePoint kernels. 
  \item[ABCD interpretability:] A modification of the previous approach that focuses on interpretability. This approach favors additive components as they are more interpretable by the practitioners. Similarly, the authors decided to remove the Rational Quadratic kernel as it is more difficult to describe automatically \cite{lloyd_automatic_2014}. 
\end{description}

All the compositions of kernels that are included in GPSS can be represented in our grammar. Thus, the search space of EvoCov is a superset of the search space of GPSS. On the other hand, ABCD approaches include ChangeWindow and ChangePoint kernels that cannot be modeled with the current grammar of EvoCov. Hence, different kernels can be found by these approaches.

\begin{figure*}[!ht]
    \centering  
    \includegraphics[width=0.9\textwidth, trim={5 30 30 50},clip]{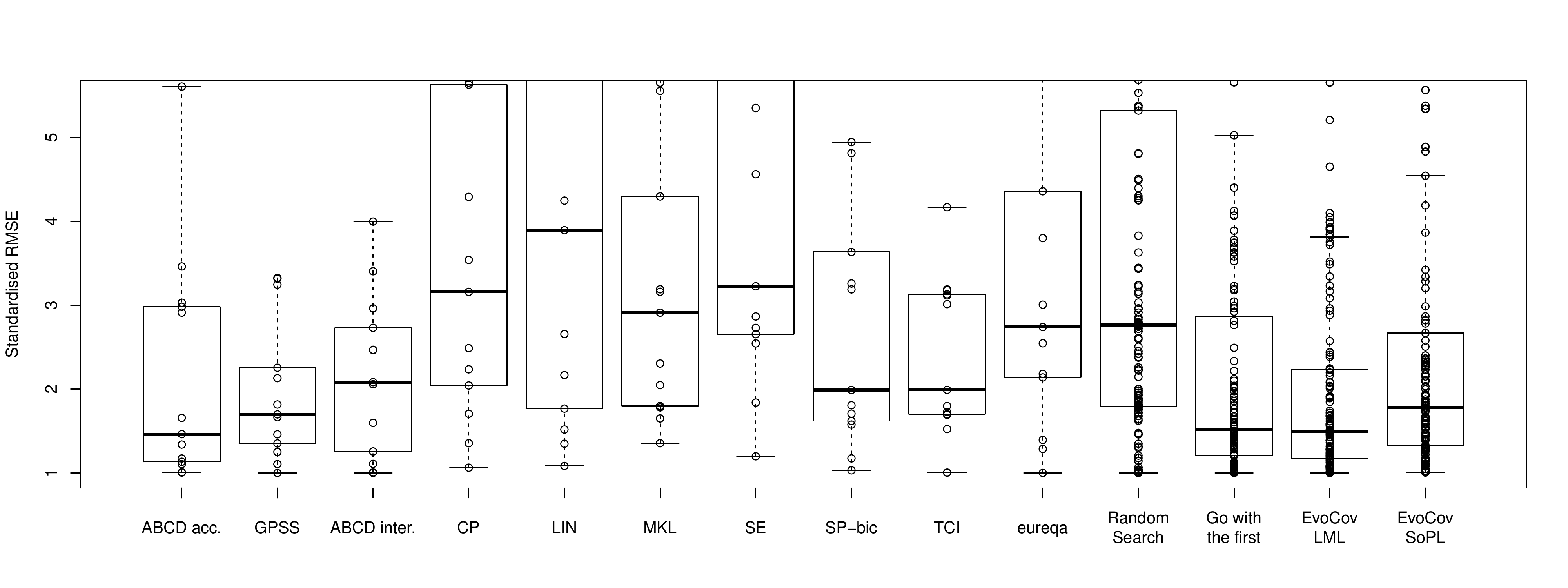}
    \caption{Standardized RMSE of each algorithm in each problem. Note that the results of our algorithms have more observations due to the 10 trials.}
    \label{fig:extrap_boxplot}
\end{figure*}

Table \ref{tab:extrap_results} shows the numerical results of the experimentation for each time-series, while in Figure \ref{fig:extrap_boxplot}, the overall results are shown. EvoCov is presented in two variants, one using LML to optimize the hyperparameters, and the other variant using the SoPL metric. Note that RMSEs are standardized by dividing by the smallest RMSE achieved in the experiments for each data set, so that the best performance on each data set has a value of 1. Also, it is worth mentioning that, in the experiments conducted in \cite{lloyd_automatic_2014}, only one trial for each time-series and algorithm was carried out\footnote{The results were gathered from the supplementary material of \cite{lloyd_automatic_2014}.}, and, for our algorithms, the mean and the best of ten trials is shown.




\begin{table*}[!ht]
  \centering
  \resizebox{\textwidth}{!}{
    \begin{tabular}{|l|rrrrrrrrrrrrrrrrrr|}
      \hline
       &     ABCD acc.                          & GPSS & ABCD intr. & CP   & LIN   & MKL   & SE   & SP-bic & TCI   & eureqa & \multicolumn{2}{c}{Random Search} & \multicolumn{2}{c}{Go With The First} & \multicolumn{2}{c}{EvoCov LML} & \multicolumn{2}{c|}{EvoCov SoPL} \\
       & & & & & & & & & & & Best & Mean & Best & Mean & Best & Mean & Best & Mean \\
      \hline
      \hline    
      Airline                         & 1.338 & 1.352 & 2.465 & 5.629 & 5.797  & 5.554 & 37.804 & \textbf{1.174}  & 1.693 & 3.799   & 1.178 & 3.311  & 1.018 & 1.446 & 1.00* & 1.307 & 1.106 & 1.559  \\
      Solar                           & 1.656 & 2.130 & 2.082 & 1.705 & 1.767  & 1.650 & 2.655  & 1.989  & 1.700 & 4.358   & 1.00* & 2.394  & 1.082 & 1.837 & 1.073 & 1.844 & 1.011 & \textbf{1.597}  \\
      Mauna Loa Atmospheric $CO_2$                       & 3.461 & 1.460 & 2.468 & 4.291 & 7.865  & 4.296 & 4.561  & 3.258  & 3.179 & 6.278   & 8.531 & 23.441 & 1.00* & 1.499 & 1.073 & \textbf{1.349} & 1.077 & 2.498  \\
      Beveridge Wheat Price Index   & 1.133 & 1.106 & 1.256 & \textbf{1.063} & 1.084  & 3.189 & 3.225  & 3.188  & 3.188 & 1.394   & 1.010 & 2.078  & 1.00* & 2.370 & 1.044 & 1.293 & 1.317 & 1.774  \\
      Daily min. temps. in Melbourne   & 1.008 & \textbf{1.00*} & 1.005 & 1.354 & 1.517  & 1.354 & 2.729  & 1.031  & 1.004 & 1.286   & 1.002 & 1.031  & 1.002 & 1.005 & 1.003 & 1.024 & 1.005 & 1.019  \\
      Internet traffic data (bits) & \textbf{1.462} & 1.698 & 2.962 & 6.099 & 7.209  & 5.976 & 6.044  & 4.944  & 3.130 & 9.149   & 5.320 & 10.779 & 1.641 & 3.893 & 1.00* & 3.626 & 1.459 & 4.824  \\
      Monthly average daily calls   & 2.981 & 2.255 & \textbf{1.00*} & 3.539 & 28.760 & 1.798 & 22.628 & 11.035 & 1.798 & 493.303 & 1.206 & 7.764  & 1.328 & 6.822 & 1.163 & 2.814 & 1.446 & 11.692 \\
      Monthly critical radio frequencies   & 5.606 & 3.316 & 3.404 & 5.652 & 7.790  & 5.652 & 15.291 & 1.806  & 4.168 & 2.545   & 1.314 & 2.687  & 1.00* & \textbf{1.346} & 1.069 & 1.682 & 1.778 & 4.077  \\
      Monthly prod. of gas in Aus.   & \textbf{1.005} & 3.325 & 2.058 & 3.159 & 2.656  & 2.910 & 2.865  & 1.620  & 1.523 & 2.739   & 1.794 & 5.017  & 1.00* & 2.754 & 1.059 & 2.499 & 1.387 & 2.340  \\
      Monthly prod. of sul. acid in Aus.   & \textbf{1.105} & 1.816 & 1.596 & 2.488 & 3.893  & 1.782 & 1.198  & 1.576  & 1.991 & 2.181   & 1.679 & 2.043  & 1.00* & 1.519 & 1.081 & 1.426 & 2.034 & 2.328  \\
      Monthly U.S. male unemployment   & 2.911 & 1.664 & 2.730 & 2.235 & \textbf{1.348}  & 2.304 & 2.544  & 4.812  & 3.012 & 3.006   & 1.315 & 5.247  & 1.299 & 1.585 & 1.00* & 1.742 & 1.168 & 1.836  \\
      Number of daily births in Quebec   & 1.173 & 1.251 & \textbf{1.108} & 2.042 & 2.167  & 2.047 & 1.839  & 1.707  & 1.725 & 2.138   & 1.300 & 1.738  & 1.00* & 1.128 & 1.049 & 1.210 & 1.135 & 1.447  \\
      Real daily wages in England ($\pounds$)   & 3.029 & 3.244 & 3.997 & 5.835 & 4.246  & 3.160 & 5.350  & 3.634  & 3.116 & \textbf{1.00*}   & 2.610 & 3.041  & 2.992 & 3.595 & 2.881 & 3.554 & 1.277 & 2.561  \\
      \hline    
      Mean                               & 2.144 & 1.971 & 2.164 & 3.469 & 5.854  & 3.206 & 8.364  & 3.213  & 2.402 & 41.014  &       & 5.428  &       & 2.362 &       & \textbf{1.951} &       & 2.998  \\
      Median                             & \textbf{1.462} & 1.698 & 2.082 & 3.159 & 3.893  & 2.910 & 3.225  & 1.989  & 1.991 & 2.739   &       & 2.684  &       & 1.559 &       & 1.478 &       & 1.831 \\
      \hline
    \end{tabular}
  }
  \vspace{0.05cm}
  \caption{Standardized RMSE for each extrapolation problem and algorithm is shown. In our approaches, the mean and the best values of the results are illustrated. The best results on average for each time-series are shown in bold, while the best results overall are highlighted by an asterisk.}
  \label{tab:extrap_results}
\end{table*}

\begin{table*}[!ht]
  \centering
  \resizebox{0.95\textwidth}{!}{
    \begin{tabular}{|l|rrrrrrrrrrrrr|}
      \hline
       &     ABCD acc.                          & GPSS & ABCD intr. & CP   & LIN   & MKL   & SE   & SP-bic & TCI   & Random Search & Go With The First & EvoCov LML & EvoCov SoPL \\
      \hline
      \hline    
      Airline                         & 12 & 15 & 12 & 8  & 4 & 5 & 3 & 16 & 8  & 6.1 & 6.3 & 10.2 & 8.5  \\
      Solar                           & 23 & 13 & 19 & 18 & 4 & 7 & 3 & 13 & 10 & 4.8 & 5.4 & 7.0  & 6.3  \\
      Mauna Loa Atmospheric $CO_2$                       & 10 & 11 & 12 & 5  & 4 & 5 & 3 & 12 & 8  & 6.7 & 4.9 & 7.0  & 10.8 \\
      Beveridge Wheat Price Index   & 14 & 7  & 13 & 13 & 4 & 5 & 3 & 7  & 5  & 5.4 & 4.9 & 6.6  & 5.8  \\
      Daily min. temps. in Melbourne   & 9  & 9  & 8  & 6  & 4 & 6 & 3 & 9  & 7  & 4.5 & 4.5 & 4.2  & 5.3  \\
      Internet traffic data (bits) & 18 & 15 & 26 & 13 & 4 & 5 & 3 & 13 & 8  & 4.1 & 6.0 & 10.1 & 6.3  \\
      Monthly average daily calls   & 18 & 19 & 16 & 11 & 4 & 7 & 3 & 7  & 7  & 5.1 & 8.2 & 11.3 & 7.6  \\
      Monthly critical radio frequencies   & 15 & 9  & 13 & 5  & 4 & 5 & 3 & 12 & 8  & 6.0 & 6.1 & 7.6  & 8.0  \\
      Monthly prod. of gas in Aus.   & 28 & 21 & 21 & 18 & 4 & 5 & 3 & 13 & 11 & 5.8 & 8.0 & 13.4 & 8.7  \\
      Monthly prod. of sul. acid in Aus.   & 19 & 17 & 17 & 13 & 4 & 7 & 3 & 12 & 9  & 5.4 & 6.5 & 7.5  & 7.2  \\
      Monthly U.S. male unemployment   & 13 & 15 & 10 & 9  & 4 & 4 & 3 & 18 & 8  & 6.4 & 4.8 & 7.1  & 8.3  \\
      Number of daily births in Quebec   &  -  & 11 & 6  & 5  & 4 & 5 & 3 & 9  & 6  & 5.0 & 5.6 & 6.8  & 7.0  \\
      Real daily wages in England ($\pounds$)   &   - & 13 & 19 & 18 & 4 & 7 & 3 & 10 & 7  & 3.0 & 5.0 & 6.3  & 5.6 \\
      \hline
      Mean & 16.3 & 13.5 & 14.8 & 10.9 & 4 & 5.6 & 3 & 11.6 & 7.8 & 5.3 & 5.9 & 8.1 & 7.3 \\
      \hline
    \end{tabular}
  }
  \vspace{0.15cm}
  \caption{Number of hyperparameters for each extrapolation problem and algorithm. The noise hyperparameter is also considered. In our approaches, the average number of hyperparameters is shown. In ABCD acc. some data is missing as it could not be found in the supplementary material of \cite{lloyd_automatic_2014}.}
  \label{tab:extrap_hp_count}
\end{table*}

As we can see in Table \ref{tab:extrap_results}, EvoCov LML achieves the best average result and beats the rest of the algorithms in the \textit{Mauna Loa Atmospheric $CO_2$} time-series. Moreover, its median result is the second best, very close to the best one. The best trials of this algorithm outperform the rest of the algorithms in three other problems. Our Go With The First algorithm is the best choice in \textit{Monthly critical radio frequencies} time-series, and its best trials were better than the other algorithms in 6 problems. Although its average performance is not as good as EvoCov LML, it was able to achieve the third best median RMSE. EvoCov SoPL shows the best behavior in \textit{Solar} time-series. On the contrary, this approach obtains poor results in \textit{Monthly average daily calls}, compromising its average result. The Random Search was able to get the best result for the \textit{Solar} time-series in its best trial. However, this algorithm has a worse performance than those already mentioned, confirming the contribution of the mutation and crossover operators. Regarding the compositional kernel search approaches, GPSS approach gets the second best mean result, and is able to beat the rest of the algorithms in \textit{Daily minimum temperatures in Melbourne} time-series. ABCD accuracy obtains the third best mean result. This algorithm gets the best median result and holds the best results in three of the time-series. ABCD interpretability, with worse results than ABCD accuracy according to the mean, is the best approach in \textit{Monthly average daily calls} and \textit{Number of daily births in Quebec} problems. On the other hand, there are some other time-series, such as \textit{Beveridge Wheat Price Index}, \textit{Monthly U.S. male unemployment} and \textit{Airline}, where the algorithms based on the CP, Linear and Spectral kernels, get better results than compositional kernel search approaches. Finally, out of the GP approaches, Eureqa outperforms the rest of the approaches in \textit{Real daily wages in England} time-series. However, this symbolic regression engine is worse than EvoCov LML in the rest of the time-series.

Table \ref{tab:extrap_hp_count} shows the number of hyperparameters of the best kernel found for each algorithm in each problem. It can be seen that the compositional kernel approaches (ABCD interpretability, ABCD accuracy and GPSS) have always more hyperparameters than any of our approaches on average. The only exception can be found in \textit{Number of daily births in Quebec} time-series, where ABCD interpretability uses 6 hyperparameters, and EvoCov LML and EvoCov SoPL use 6.8 and 7.0 hyperparameters respectively.

Overall, EvoCov LML is a competitive approach compared to the current state-of-the-art compositional kernel search approaches. In spite of not including the ChangePoint and ChangeWindow kernels, this approach is able to obtain comparable results to ABCD accuracy, with kernels that have fewer hyperparameters, which makes them easier to optimize.

\subsection{Comparing our proposal to ad hoc kernel approaches}
\label{ssec:vsadhoc}

As we have mentioned in Section \ref{sec:relwork}, many works in GPs propose a human-designed specific kernel for each particular problem. One of the most recent works can be found in \cite{preotiuc-pietro_temporal_2013}, where the number of tweets in the Twitter timeline that contain a given hashtag was predicted by means of a GP. The authors show that this information can be useful to predict the hashtags that a tweet has, given its content. They propose an ad hoc kernel, Periodic Spikes (PS), that captures the periodicities of these hashtag time-series. For example, the \textit{\#goodmorning} hashtag shows a clear periodic pattern, as it is more frequently tweeted in the mornings. On the other hand, there are some hashtags, such as \#np (now playing), that do not follow the periodic pattern mentioned above, and according to the authors, there are kernels better suited than PS for these problems. Our proposal should be able to identify these situations, and offer the best possible kernel without human intervention.

Hence, we carried out the same experiment as in \cite{preotiuc-pietro_temporal_2013}, where \textit{\#goodmorning}, \textit{\#breakfast}, \textit{\#confessionhour}, \textit{\#fail}, \textit{\#fyi} and \textit{\#raw} hashtags are predicted\footnote{Data can be found in \url{https://web.sas.upenn.edu/danielpr/resources/}}. For each hashtag, the number of tweets per hour was collected, using one month for training and the other to test, except for the \textit{\#goodmorning} hashtag, as in the original paper, where 3 weeks were gathered, having 2 weeks to train and the last one to test.

\begin{figure}[!ht]
    \centering   
    \includegraphics[width=0.6\textwidth,trim={0 10 0 5},clip]{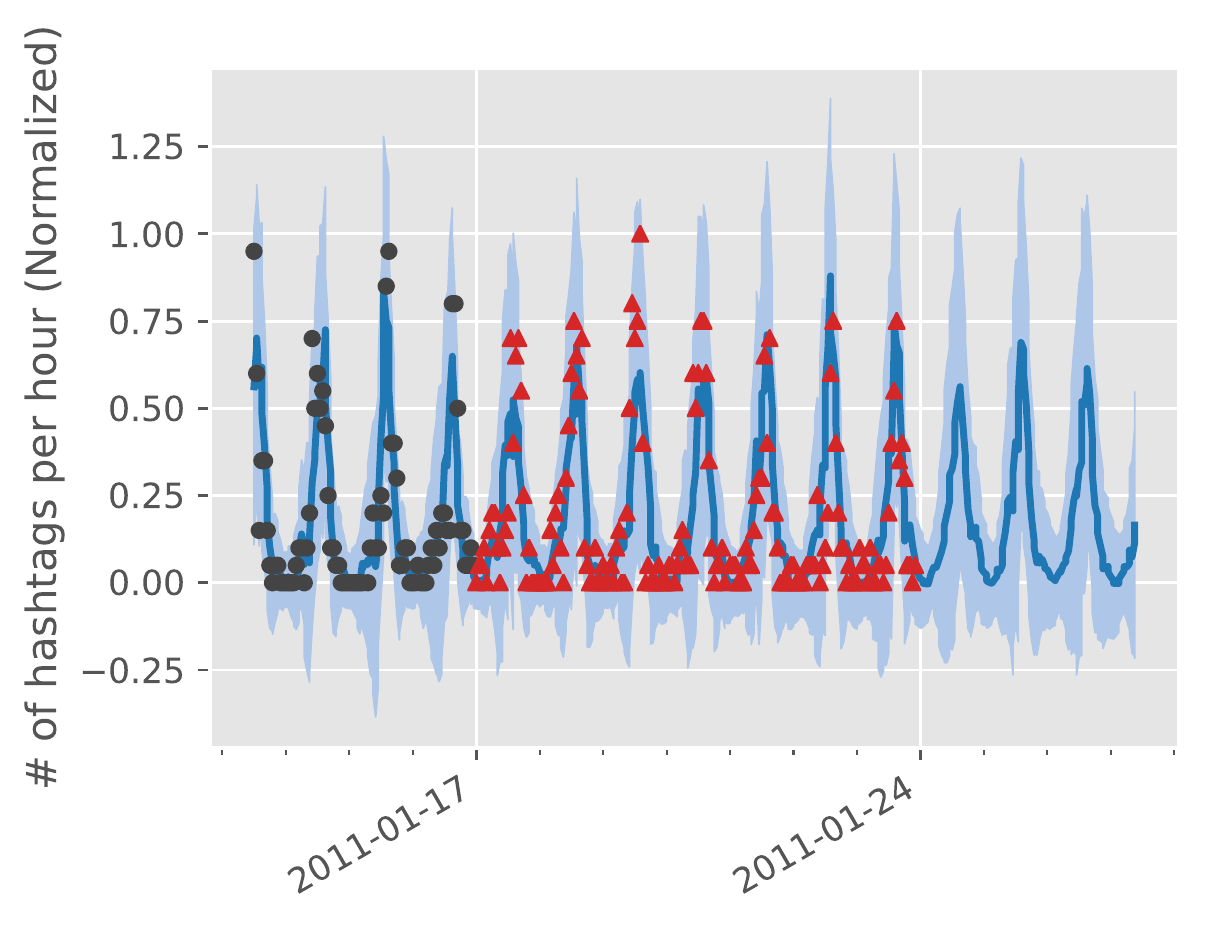}
    \caption{Extrapolation of \textit{\#goodmorning} hashtag time-series. The black dots represent the last samples of the train set, while the red triangles show the samples of the test set. The prediction given by a GP with a kernel learned by the EvoCov LML method is illustrated with a continuous blue curve for the mean and the light blue shadow shows 3 times the standard deviation.}
    \label{fig:hashtag_goodmorning3week}  
\end{figure}

In Figure~\ref{fig:hashtag_goodmorning3week}, an example of the hashtag prediction is shown. The number of occurrences of the \textit{\#goodmorning} hashtag is shown per hour, along with the best model given by our approach in a single run. In this problem, a periodic trend can be appreciated, which is successfully captured by our model.

Table~\ref{tab:hashtag_results} shows a comparison between EvoCov LML, EvoCov SoPL and the PS kernel. The experiments with the PS kernel were carried out using our software, and $ref\_fun\_call=5000$ samples were allowed to find the best hyperparameters for this kernel. As can be seen, EvoCov SoPL is the best approach on average, and it obtains the best results in the \textit{\#confessionhour} problem. EvoCov LML is able to get the best score in the \textit{\#fail} and \textit{\#fyi} hashtags, finding a complex periodic pattern. It is also worth mentioning that the best trials of this algorithm obtain the best results in four out of five problems. However, in simpler periodic time-series, such as \textit{\#goodmorning}, \textit{\#breakfast} and \textit{\#raw}, the PS kernel is the best choice, getting the best median result.

\begin{table}[!ht]
  \centering
  \begin{tabular}{|l|rrrrr|}
    \hline
    & PS & \multicolumn{2}{c}{EvoCov LML}   & \multicolumn{2}{c|}{EvoCov SoPL} \\
    & & Best & Mean & Best & Mean \\
    \hline
    \hline        
    \#goodmorning      & \textbf{1.072}   & 1.00* & 1.074   & 1.374 & 1.528   \\
    \#breakfast        & \textbf{1.014}   & 1.00* & 1.085   & 1.021 & 1.021   \\
    \#confessionhour   & 323.949 & 1.048 & 220.262 & 1.00* & \textbf{147.540} \\
    \#fail             & 1.028   & 1.00* & \textbf{1.018}   & 1.003 & 1.019   \\
    \#fyi              & 1.051   & 1.00* & \textbf{1.012}   & 1.031 & 1.045   \\
    \#raw              & \textbf{1.159}   & 1.00* & 1.407   & 1.543 & 1.554   \\
    \hline
    Mean                      & 54.879  &       & 33.989  &       & \textbf{23.069}  \\
    Median                    & \textbf{1.062}   &       & 1.077   &       & 1.236   \\
    \hline
  \end{tabular}
  \vspace{0.15cm}
  \caption{The PS ad hoc kernel compared to EvoCov LML and EvoCov SoPL, in hashtag prediction problems. Standardized RMSE for each extrapolation problem and algorithm is shown. In our approaches, the mean and the best results are illustrated. The best results on average are shown in bold, while the best results overall are highlighted by an asterisk.}
  \label{tab:hashtag_results}
\end{table}

Table~\ref{tab:hashtag_hp_count} shows the number of hyperparameters used by different approaches. As expected, our approaches use more hyperparameters than the PS kernel, as this kernel is specifically designed for these problems.

\begin{table}[!ht]
  \centering
  \begin{tabular}{|l|rrr|}
    \hline
    & PS & EvoCov LML   & EvoCov SoPL \\
    \hline
    \hline        
    \#goodmorning & 3 & 7.9  & 7.9 \\
    \#breakfast        & 3 & 6.2  & 9.0 \\
    \#confessionhour   & 3 & 14.4 & 7.8 \\
    \#fail             & 3 & 6.2  & 5.0 \\
    \#fyi              & 3 & 3.1  & 4.3 \\
    \#raw              & 3 & 12.6 & 6.8 \\
    \hline
    Mean & 3 & 8.4 & 6.8 \\
    \hline
  \end{tabular}
  \vspace{0.15cm}
  \caption{Number of hyperparameters for each extrapolation problem and algorithm. The noise hyperparameter is also considered. In our approaches, the average number of hyperparameters is shown.}
  \label{tab:hashtag_hp_count}
\end{table}

The PS kernel is still able to hold the best results in three out of six problems. On the other hand, the EvoCov approaches, without any expertise about the problem, are able to obtain similar predictions to PS, even improving the results of the PS kernel in the rest of the problems. Also note that the EvoCov approaches are able to get better average results than the PS kernel, showing a more adaptable behavior.









\section{Conclusions}
\label{sec:conclu}

Kernel functions are widely used in several Machine Learning methods. GPs are one of these techniques, where a PSD kernel is used as a covariance function. This kernel function has to be carefully selected to achieve good results in any GP application. Although initial approaches used to rely on predefined kernels or ad hoc solutions for specific problems, there is an increasing interest in automatically learning these kernels. In this work, we have presented an evolutionary approach to learn kernel functions for GPs. While other approaches are based on kernel composition, in our approach, kernels are modeled by means of basic mathematical expressions.

This work has made the following contributions:

\begin{itemize}
  \item Basic mathematical expressions as building blocks for GP kernels: We propose to bring the progress made in other Machine Learning areas to the GPs by considering its covariance function as a program that can be learned.
  \item Fast PSD check for GP kernels: Although some of the kernels generated by this new random method are not PSD, we have defined a kernel validation procedure that rapidly discards most non-PSD expression trees based on the properties of the covariance matrix.
  \item Hyperparameter inheritance: We have incorporated hyperparameter inheritance within GenProg, improving the efficiency of the algorithm.
  \item Metric comparison for hyperparameter optimization: We provide valuable insights about the suitability and performance of several metrics for hyperparameter optimization in extrapolation problems.
  \item Extensive benchmark in realistic problems: We have evaluated our proposal in an extensive benchmark of realistic problems, showing that our agnostic algorithm is competitive to a wide range of methods.
\end{itemize}

Altogether, these contributions enabled the design of a GenProg variant which is able to improve the state-of-the-art results in the application of GP to time-series extrapolation. We can conclude that there is no need to rely on a priori defined kernels for GP time-series extrapolation problems, as it is possible to learn simpler and better kernels by evolving mathematical expression trees that satisfy the PSD restrictions.

Further research in the grammar is suggested, extending it to ChangePoint and ChangeWindow kernels. On the other hand, we propose continuing the work carried out to measure the performance of the hyperparameter optimization metrics for GP extrapolation problems.

\section*{Acknowledgments}
This work has been partially supported by the Basque Government (IT1244-19 and ELKARTEK programs), and Spanish Ministry of Economy and Competitiveness MINECO (project TIN2016-78365-R). Jose A. Lozano is also supported by BERC 2018-2021 (Basque government) and Severo Ochoa Program SEV-2017-0718 (Spanish Ministry of Economy and Competitiveness), while Ibai Roman has held a predoctoral grant from the Basque Government.

\bibliographystyle{unsrt}  
\bibliography{biblio}

\end{document}